\documentclass[man,12pt, donotrepeattitle, letterpaper, floatsintext]{apa7} 
\usepackage[top=1in,right=1in,left=1in,bottom=1in]{geometry}
\usepackage{amsmath}
\usepackage{booktabs}
\usepackage{adjustbox}
\usepackage{longtable}
\usepackage{float}
\usepackage{multicol}
\usepackage{caption, subcaption}

\usepackage{etoolbox} 

\usepackage[style=apa, natbib=true, backend=biber, sortcites=true, sorting=nyt, autocite=inline]{biblatex}
\DeclareLanguageMapping{american}{american-apa}
\usepackage[toc,page]{appendix}

\usepackage{graphicx,tabularx,url,multirow,setspace,caption, comment, amsmath}

\title{Prediction of Item Difficulty for Reading Comprehension Items\\by Creation of Annotated Item Repository}
\shorttitle{Prediction of Item Difficulty}
\authorsnames{Radhika Kapoor, Sang T. Truong,\\Nick Haber, Maria Araceli Ruiz-Primo, Benjamin W. Domingue}
\authorsaffiliations{Stanford University}
\date{\today}
\leftheader{}
\abstract{\noindent Prediction of item difficulty based on its text content is of substantial interest. In this paper, we focus on the related problem of recovering IRT-based difficulty when the data originally reported item p-value (percent correct responses). We model this item difficulty using a repository of reading passages and student data from US standardized tests from New York and Texas for grades 3-8 spanning the years 2018-23. This repository is annotated with meta-data on (1) linguistic features of the reading items, (2) test features of the passage, and (3) context features. A penalized regression prediction model with all these features can predict item difficulty with RMSE 0.59 compared to baseline RMSE of 0.92, and with a correlation of 0.77 between true and predicted difficulty. We supplement these features with embeddings from LLMs (ModernBERT, BERT, and LlAMA), which marginally improve item difficulty prediction. When models use only item linguistic features or LLM embeddings, prediction performance is similar, which suggests that only one of these feature categories may be required. This item difficulty prediction model can be used to filter and categorize reading items and will be made publicly available for use by other stakeholders.}


\authornote{Correspondence concerning this article should be sent to Radhika Kapoor, Stanford University, email:  rkap786@stanford.edu. This work was funded in part by Stanford Data Science
}

\begin{document}
\maketitle
\setcounter{secnumdepth}{3}
\section{Introduction}
Education systems use reading comprehension tests to evaluate student achievement and growth, to monitor average school and teacher performance, and to diagnose students for learning disorders. Creating items for these tests can be an expensive and time-intensive endeavor \citep{case1998constructing, rudner2009implementing}, especially for standardized tests in K-12 schools. Conventionally, subject matter experts (e.g., teachers or reading specialists) create the items used in standardized tests \citep{attali2014estimating}. Items are then piloted in schools and student response data is collected to estimate item properties like difficulty and to confirm that items are measuring student knowledge as intended \citep[Chapter~1]{lane2016handbook}.\footnote{New York: https://www.nysed.gov/state-assessment/nysed-test-development-process; Texas: https://tea.texas.gov/texas-schools/accountability/academic-accountability/performance-reporting/8-why-does-texas-field-test.pdf} However, items developed by subject matter experts might not perform as expected when taken to the field \citep{bejar2012item, reid1991training, impara1998teachers, dee2021assessing}. 

Education researchers have long been interested in modeling the difficulty of test items. Improved item difficulty models could support subject matter experts in creating items as well as reduce field testing time and costs \citep{bejar1991generative, gorin2006item}. However, this field has achieved mixed results, as modeling and predicting item difficulty is complex, and requires advanced text analysis and annotation to extract relevant item features \citep{gorin2006item}. Item difficulty is determined by the interplay of an item, the respondent, and the respondent's context \citep{beck1997using,toyama2021makes, benedetto2023survey}. For reading comprehension, item features that affect difficulty include characteristics of the reading passage (e.g., semantic or syntactic complexity and theme), characteristics of the questions asked about the passage (e.g., cognitive functioning elicited by the item or item format like multiple choice or constructed response), test characteristics (e.g., directions or instructions for the test and including reading aids like vocabulary definitions), or how these different characteristics interact with each other (e.g., how does an item relate to the reading passage?) \citep{abadiano2003rand, toyama2021makes, bulut2023evaluating, laverghetta2023generating, huang2017question}. Relevant respondent features include not only their reading ability but also their contextual features such as school or home contexts (e.g., if the respondents are familiar with the topic or passage theme or if they speak the test language at home)  


Large Language Models (LLMs) have recently shown the potential to generate items with desirable characteristics like difficulty and to automate scoring of text-based responses \citep{ferrara2018item, huang2017question, white2022automated, attali2022interactive, zelikman2023generating} possibly reducing the time and cost associated with these tasks \citep{acharya2023llm}. In this paper, we leverage LLMs with item and respondent features to predict mean difficulty of reading comprehension items. We build a repository of items using reading comprehension items from New York and Texas, administered to Grades 3-8 students spanning 5 years between 2018 and 2023. This repository contains the item text (reading passage, question, and distractors), and the state-level average percent correct responses (p-value). Items are annotated with (1) linguistic features of the reading passage, including their lexical, semantic, and syntactic characteristics and readability metrics (2) test features of the item, such as the presence of highlighted or bold text (3) respondent's state, grade, and year (4) LLM-generated text embeddings for reading comprehension items. 

We focus on how effectively LLM-generated embeddings predict item difficulty in an attempt to understand whether transformer-based models can augment traditional psychometric methods used to generate items. Our models achieve an RMSE of 0.59 compared to a baseline of 0.92, and a correlation of 0.77 between true and predicted item difficulty. Previous results for predicting difficulty using item content achieved correlation of 0.3-0.7, and smaller RMSE improvements \citep{ha2019predicting, huang2017question}. We make the following contributions to literature: First, to our knowledge, this is the first publicly available work to examine automatic difficulty modeling for reading items for standardized tests in US schools. Second, we also convert mean p-value of items from different grades to a vertical logit scale using a Rasch IRT model, which could be used in future work to aggregate response data collected from different populations. Third, we compare difficulty prediction from different types of item features: linguistic features, test features, pre-trained LLM embeddings, and respondent context. We report similar results for prediction from linguistic features or LLM embeddings, suggesting that any of these classes of metrics could be used by themselves for prediction. Fourth, we compare the performance of three LLMs (BERT-base, ModernBERT, and LlAMA), as well as different prompts for generating LLM embeddings. We report similar results for these variations.

The paper is structured as follows: the Background section reviews the literature for reading comprehension tests and for item difficulty modeling. The Data section describes the reading comprehension items from New York (NY) and Texas that were administered to students from Grades 3-8. Next, the Methods section details how we process the dataset, including adjusting p-values so they are placed on a vertical logit scale, annotating reading passages and items, and generating LLM-based embeddings. The regression penalized model used to predict difficulty using these features is also discussed. The Results section reports item difficulty modeling performance. 

\section{Background}
\subsection{Features of Reading Comprehension Items}
A reading comprehension item in a test is usually structured as a passage that the respondents read, followed by a question or prompt, and then by several options from which respondents select one response. The item might also include a few other components: directions about the passage or the item (e.g., ``read the passage and answer the questions that follow''), images or figures (e.g., to highlight part of the story), and tables (e.g., with data related to the story). This section describes the features of reading comprehension items that could predict their difficulty.

\subsubsection{Features of Reading Passages}
Text complexity of reading passages has been captured using (1) descriptive metrics, such as the count of words in passage, number of sentences, and number of paragraphs (2) linguistic complexity, which is influenced by factors such as complexity of words (measures include word length and word frequency), syntactic complexity (measures include modifier propositional density or count of adjectives per 1000 words), and text cohesion (defined as words and ideas overlapping throughout the text) \citep{graesser2014coh,mcnamara2014automated}.

Existing measures of reading levels like Lexile scores and Flesch-Kincaid metrics use a combination of these metrics to calculate the reading level of texts.  Flesch-Kincaid scores are a weighted aggregate of sentence length (number of words per sentence) and word complexity (number of syllables per word), and they range between 0 to 18. Lexile scores are a measure of sentence length and word frequency. They are used for benchmarking text at grade level such that 75\% of students in the grade can read the given text. For instance, Lexile scores are defined so that they range from 110-430L for Kindergarten to 910-1230L for Grade 6. 

Reading standards such as the Common Core Standards for English Language Arts \& Literacy\footnote{https://achievethecore.org/page/2725/text-complexity} are tied to these complexity measures, both for teaching and tests. For instance, the New York State Quantitative Text Complexity Chart for Assessment and Curriculum\footnote{https://www.nysedregents.org/engageny/hsela/regela-d-3-text-complexity-form.pdf, last updated September 15, 2022} specifies that passages selected for grades 2-3 should have a Flesch-Kincaid score between 1.98-5.34 and a score between 420-820 on the Lexile framework; grades 4-5 should be between 4.97-7.03 on Flesch-Kincaid and between 740-1010 on Lexile framework; grades 6-8 should be between 6.51-10.34 on Flesch-Kincaid and between 925-1185 on Lexile scales. 


Corpus analysis tools can parse text for their linguistic features. A popular corpus analysis tool is the Coh-Metrix, which offers over 100 measures of vocabulary, syntax, semantics, and other linguistic features. Coh-Metrix also reports aggregate measures of text cohesion and readability using metrics like Flesch-Kincaid, latent semantic analysis and principal components analysis. \citep{mcnamara2014automated}. Measures from corpus analysis tools, such as length of passage, concreteness or abstractness of text, and level of vocabulary have been reported to be correlated with difficulty for reading questions \citep{alkhuzaey2023text,choi2020predicting}. 

In addition to the reading complexity of the passage, its layout or structure might also affect student performance. The layout here refers to reading cues that point students to important aspects of the content (such as text segmentation), the inclusion of boxes or footnotes, directions that point readers to important vocabulary, helpful images, and options for text-to-speech in computer-based reading \citep{lawrence2022reading, mize2020developing}. 

\subsubsection{Features of Items}
Item features affect their difficulty by changing their interpretability. Some features can make it easier to locate and retrieve information from text \citep{lawrence2022reading, le2017sources, ruizprimo2015}. For example, an item can include relevant sentences from the passage in a text box or definitions of difficult words. Other features emphasize key aspects of items. For instance, the word ``agree'' in the item is emphasized: ``Which of the following statements would <name of character> in the story \textbf{agree} with?''. In multiple choice questions, the distractors or response options also affect difficulty; for instance, some distractor options could be very similar to the correct answer. The type of content (e.g., use of images) might also influence student performance.



\subsection{Item Difficulty Modeling}
\subsubsection{Automatic generation of items}
Automated item generation was first proposed in the 1970s \citep{gierl2013automatic}, though the field saw a resurgence in the 2000s with the need for more item creation with computerized tests, as well as with an increase in computational power. A type of automatic item generation uses expert-created templates, item schemes or structures, parts of which can be populated using algorithms \citep{embretson200623, kurdi2020systematic}. Such cognitive item model templates are specific to the item type being created.  Item templates can define: (1) structure of an item (2) cognitive processes elicited by an item (3) item features that manipulate or affect desired cognitive processes. For example, in math, a template for double-digit addition might allow a range of numbers; in a medical exam item, the gender or age of the patient can be generated by the algorithm \citep{karamanis2006generating}. Other approaches to creating items use manipulation using semantic or syntactic rules to create items e.g., question sentences and distractors can be created based on sentence similarity, where similarity can be defined using a cosine index.


With recent advances, LLMs can be prompted to generate items without additional tuning or information (``zero-shot'' generation) or with some examples in context (``one-shot'' or ``few-shot'' generation) \citep{attali2022interactive,kurdi2020systematic}. LLMs can also be prompted using cognitive models, templates, or rules to create items with desired properties \citep{gierl2013automatic, kurdi2020systematic}. These templates can specify the type of item to be generated (e.g., fill-in-the-blank questions, multiple-choice questions, open-ended questions), provide further instructions based on item type (e.g., the number of distractors and correct responses for multiple choice items), specify reasoning to be elicited (e.g., recall or inference), and even state the desired item difficulty (e.g., ``create an easy item that 70\% of respondents would get right'')  \citep{sayin2024using}. 



\subsubsection{Item difficulty modeling}
The problem of item generation is fundamentally linked with the problem of item difficulty modeling. Items are created with an implicit or explicit difficulty level, defined by human experts using heuristics or by statistical modeling. Popular models used for statistical modeling include supervised learning approaches such as linear regression, decision trees, and random forests. Unsupervised learning approaches like deep learning and neural nets have also been leveraged but are less common, likely because of the smaller size of education datasets and the desire for inference in addition to model prediction. 


Statistical item difficulty modeling approaches usually have the following components: (1) A method to parse item text and images: this can use NLP and image parsers, linguistic corpus analysis tools, and human annotation (2) Model to predict difficulty. Recently, difficulty modeling has been supplemented by embeddings from LLMs \citep{kurdi2020systematic,sharpnack2024banditcat}. Embeddings can be extracted from pre-trained models, or models could be fine-tuned. In the shared task organized during the BEA workshop at NAACL 2024 \citep{yaneva2024findings}, item difficulty (defined as the percentage of respondents who responded incorrectly)  was modeled for 667 MCQ items from the United States Medical Licensing Examination. The best prediction model achieved RMSE of 0.29 compared to a baseline RMSE of 0.31 from a dummy regressor model; the best model used a combination of LLM embeddings and other item features. In a similar exercise with 12,038 MCQ items from the United States Medical Licensing Examination, the best model used a combination of embeddings from Word2Vec and ElMo and linguistic features together, achieving RMSE of 22.45 compared to a baseline RMSE of 23.65. 

The results of existing item difficulty prediction efforts have achieved limited success. A key constraint is that education settings rarely provide sufficiently large datasets for fine-tuning LLMs. This paper considers if pre-trained LLMs can add value for predicting the difficulty of reading comprehension items. The next section describes the repository of items used in the paper, which has been annotated with relevant item characteristics such as reading passage complexity and available reading aids. This will be used along with LLM-generated embeddings to predict item difficulty.

\section{Data}
The dataset contains multiple-choice items from the New York State Testing Program (NYSTP) and Texas STAAR reading comprehension standardized tests for Grades 3-8. NYSTP data is available for the years 2018, 2019, and 2022, and Texas STAAR data is available for 2019, 2021, and 2022. In total, there are 1076 items based on 170 passages across grades and years from both states (Table \ref{tab:grade_data_totals}).

\begin{table}[htb]
\centering
\caption{Count of Items for Grades by State and Year}
    \begin{tabular}{cccccccc}
    \toprule
    \multirow{2}{*}{\textbf{State}} & \multirow{2}{*}{\textbf{Grade}} & \multicolumn{5}{c}{\textbf{Item counts by year}} & \multirow{2}{*}{\textbf{Total}} \\ \cmidrule(lr){3-7}
    &  & \textbf{2018} & \textbf{2019} & \textbf{2021} & \textbf{2022} & \textbf{2023} &  \\ \midrule
    NY & Grade 3 & 12 & 12 &  & 12 & 17 & 53 \\ 
    NY & Grade 4 & 12 & 12 &  & 12 & 17 & 53 \\ 
    NY & Grade 5 & 21 & 21 &  & 21 & 19 & 82 \\ 
    NY & Grade 6 & 21 & 21 &  & 21 & 19 & 82 \\ 
    NY & Grade 7 & 21 & 21 &  & 21 & 26 & 89 \\ 
    NY & Grade 8 & 21 & 21 &  & 21 & 26 & 89 \\ 
    Texas & Grade 3 &  & 34 & 28 & 34 &  & 62 \\ 
    Texas & Grade 4 &  & 32 & 32 & 32 &  & 96 \\ 
    Texas & Grade 5 &  & 29 & 34 & 34 &  & 97 \\ 
    Texas & Grade 6 &  & 36 & 36 & 36 &  & 108 \\ 
    Texas & Grade 7 &  & 37 & 37 & 38 &  & 112 \\ 
    Texas & Grade 8 &  & 40 & 39 & 40 &  & 119 \\ \hline
    Total &  & 108 & 316 & 206 & 322 & 124 & 1076 \\ \hline
    \end{tabular}
\label{tab:grade_data_totals}
\end{table}

\begin{table}[htb]
\centering
\caption{Item p-Values by Grades, State and Year}
    \begin{tabular}{ccccccc}
    \toprule
    \multirow{2}{*}{\textbf{State}} & \multirow{2}{*}{\textbf{Grade}} & \multicolumn{5}{c}{\textbf{Item counts by year}}  \\ \cmidrule(lr){3-7}
    &  & \textbf{2018} & \textbf{2019} & \textbf{2021} & \textbf{2022} & \textbf{2023}   \\ \midrule
    NY & Grade3 & 0.60 & 0.65 &  & 0.65 & 0.60 \\ 
    NY & Grade4 & 0.58 & 0.57 &  & 0.64 & 0.56 \\ 
    NY & Grade5 & 0.62 & 0.64 &  & 0.65 & 0.60  \\ 
    NY & Grade6 & 0.67 & 0.65 &  & 0.67 & 0.58  \\ 
    NY & Grade7 & 0.55 & 0.60 &  & 0.66 & 0.59  \\ 
    NY & Grade8 & 0.70 & 0.62 &  & 0.64 & 0.61  \\ 
    Texas & Grade3 &  & 0.68 & 0.62 & 0.68    \\ 
    Texas & Grade4 &  & 0.66 & 0.62 & 0.70    \\ 
    Texas & Grade5 &  & 0.70 & 0.68 & 0.74    \\ 
    Texas & Grade6 &  & 0.64 & 0.64 & 0.65    \\ 
    Texas & Grade7 &  & 0.67 & 0.65 & 0.70    \\ 
    Texas & Grade8 &  & 0.70 & 0.67 & 0.71    \\ \hline
    \end{tabular}
\label{tab:grade_yr_pvalue}
\end{table}

Passages and related multiple-choice items are annotated with additional features. These include item difficulty, defined as mean percent correct response at the state, year, and grade level. Table \ref{tab:grade_yr_pvalue} reports mean percent correct responses (i.e., the item's \textit{p-value}\footnote{Discussed in Chapter 14 in \cite{crocker1986introduction}}) across items by grade, state, and year.  Note that mean accuracy for both states centers around 60\% across grades and years. Student acquisition of additional reading comprehension skills is being offset by increases in overall item difficulty. As a consequence, reported difficulties or p-values cannot be directly compared across grades. In other words, let us say there is a Grade 3 item with p-value 0.60 (i.e., 60\% of 3rd graders get this item right) and a Grade 8 item with p-value 0.60. The Grade 3 item is presumably easier than the Grade 8 item, since Grade 8 students are likely to be better readers; however, the p-values reported in the dataset would consider these items to be equally difficult. We convert these p-values to a common vertical logit scale to make them comparable for difficulty modeling. This is described in more detail in the Methods section.

\section{Methods}
This section describes how the dataset is processed for item difficulty prediction. Section 4.1 describes how mean p-values across grades and years are rescaled to a uniform logit vertical scale. Section 4.2 describes annotation of items with linguistic, test, and context features used as inputs in the prediction model. Section 4.3 describes how embeddings are generated from pre-trained LLMs. Finally, Section 4.4 then describes how these inputs are used to predict item difficulty using a penalized regression model.

\subsection{Rescaling p-values}

We use publicly available estimates of average student abilities across grades to convert average p-value to a common vertical scale. Table \ref{tab:nwea_staar} shows the NWEA MAP 2020 reading scale\footnote{https://www.nwea.org/uploads/2020/02/NY-MAP-Growth-Linking-Study-Report-2020-07-22.pdf}, which reports mean grade-level achievement norms for over 500,000 students attending public schools across 50 states. These average scores are used to transform the p-values from a specific grade to a scale that can then be used for direct comparison; for example, on our derived scale the 3rd grade item with a p-value of 0.6 will have a lower value than the 8th grade item with a p-value of 0.6 given the growth from 200.74 to 220.93 observed on the NWEA scale. 


\vspace*{5mm}
\begin{table}[tb]
    \centering
    \caption{NWEA Spring 2020 Reading Student Achievement Norms by Grade}
        \begin{tabular}{cc}
        \hline
        Grades & Score \\ 
        \hline
        Grade 3 & 200.74 \\ 
        Grade 4 & 204.83 \\ 
        Grade 5 & 210.19  \\ 
        Grade 6 & 215.36 \\ 
        Grade 7 & 216.81 \\ 
        Grade 8 & 220.93 \\ \hline
        \end{tabular}
    \label{tab:nwea_staar}
\end{table}

\subsubsection{Rescale p-values to IRT difficulty}
To rescale the p-values to the logit scale, we take advantage of a basic feature of 1PL IRT models \citep{von2016rasch} that suggests a relationship between such p-values and item-person characteristics: 
\begin{equation}
    \Pr(y=1)= p_{ij} = \sigma (\theta_i+b_j)
    \label{irt1pl}
\end{equation}
where $\theta$ is the relevant person ability, $b$ is the item easiness, and $\sigma$ is a sigmoid function (the logistic sigmoid, $\sigma(x)=(1+\text{exp}(-x)$, which we also use here). We don't have access to the relevant $\theta_i$ and $b$ values that generated the data here, but we utilize the fact that they---in particular average $\theta$ values by grade level---are publicly reported for similar groups. We thus do the following. We denote the mean ability for students in a given grade (using values in Table \ref{tab:nwea_staar}) as $\overline{\theta_g}$. For each item $j$ given to students in grade $g$, we can calculate easiness $b_j$ based on the relationship between the p-value $p_{j}$ and the above Equation \ref{irt1pl}. Specifically, we solve for $b_j$ such that 

    \begin{align}
    p_{ij} =\frac{1}{1+\exp(-(\overline{\theta_g}+b_j))}\\
    \iff b_j = -\overline{\theta_g} + \log(\frac{p_{j}}{1-p_{j}})
    \label{eqn:diff}
    \end{align}
We use the resulting $b_j$ values in our subsequent modeling.

\begin{figure}[htb!]
    \centering
    \caption{Conversion of p-value to IRT Easiness by Grades}
    \label{pvalueconvert}
\includegraphics[width=\textwidth]{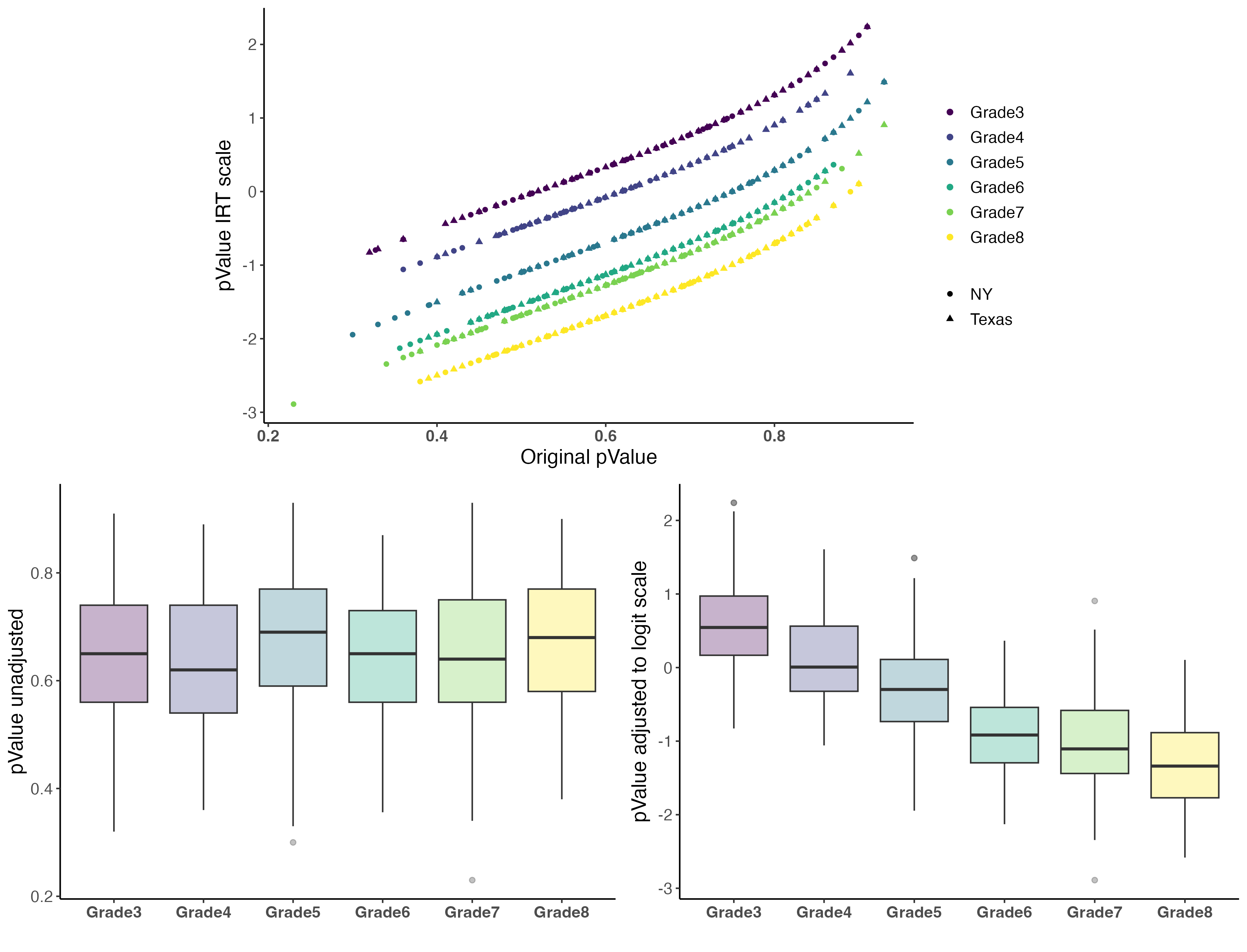}
\caption*{Note. (Top): Conversion of pValue to IRT scale by Grade; (Bottom Left): Unadjusted p-values by Grades (Bottom Right): Adjusted Mean Easiness by Grades on the Logit Scale}
\end{figure}

\subsubsection{Adjustment Results}
In the analysis below, $\overline{\theta_g}$ is drawn from the NWEA Spring 2020 reading vertical scale. Figure \ref{pvalueconvert} shows the results of the adjustment process. The top panel shows grade-level mean $b_ j$ values for NY and Texas, mapped from p-values using $\overline{\theta_g}$ using NWEA 2020 reading achievement scores. The bottom panel shows the distribution of the unadjusted and adjusted p-values by grades. The unadjusted mean p-values (bottom left panel) are all distributed around 0.6 and there is no increasing pattern as grade levels increase. This pattern is adjusted on the logit scale in the bottom right scale. Note that adjusted Grade 3 items are easiest on average, while Grade 8 items are hardest. Hence, average p-value of 0.6 for Grade 3 now corresponds to 0.3 on the adjusted scale, and maps to -1.69 for Grade 8.

Note one critical fact. The values on the y-axis in Figure \ref{pvalueconvert} are based on the NWEA Scale in Table \ref{tab:nwea_staar}. While the NWEA scale is a high-quality scale derived from large samples, it is one of several scales that could provide a common metric across grades \citep{dadey2012meta}. Given that this choice of $\overline{\theta_g}$ is somewhat arbitrary, we consider a range of potentially plausible $\overline{\theta_g}$ values in ancillary robustness analyses (discussed in Appendix \ref{Appendix B}).

\subsection{Annotation of item bank}

\begin{figure}[htb!]
    \centering
    \caption{Predictors for Item Difficulty Model: Input text to generate embeddings, and Linguistic, Test, and Context Features}
    \label{EmbeddingGen}
\includegraphics[width=\textwidth,keepaspectratio]{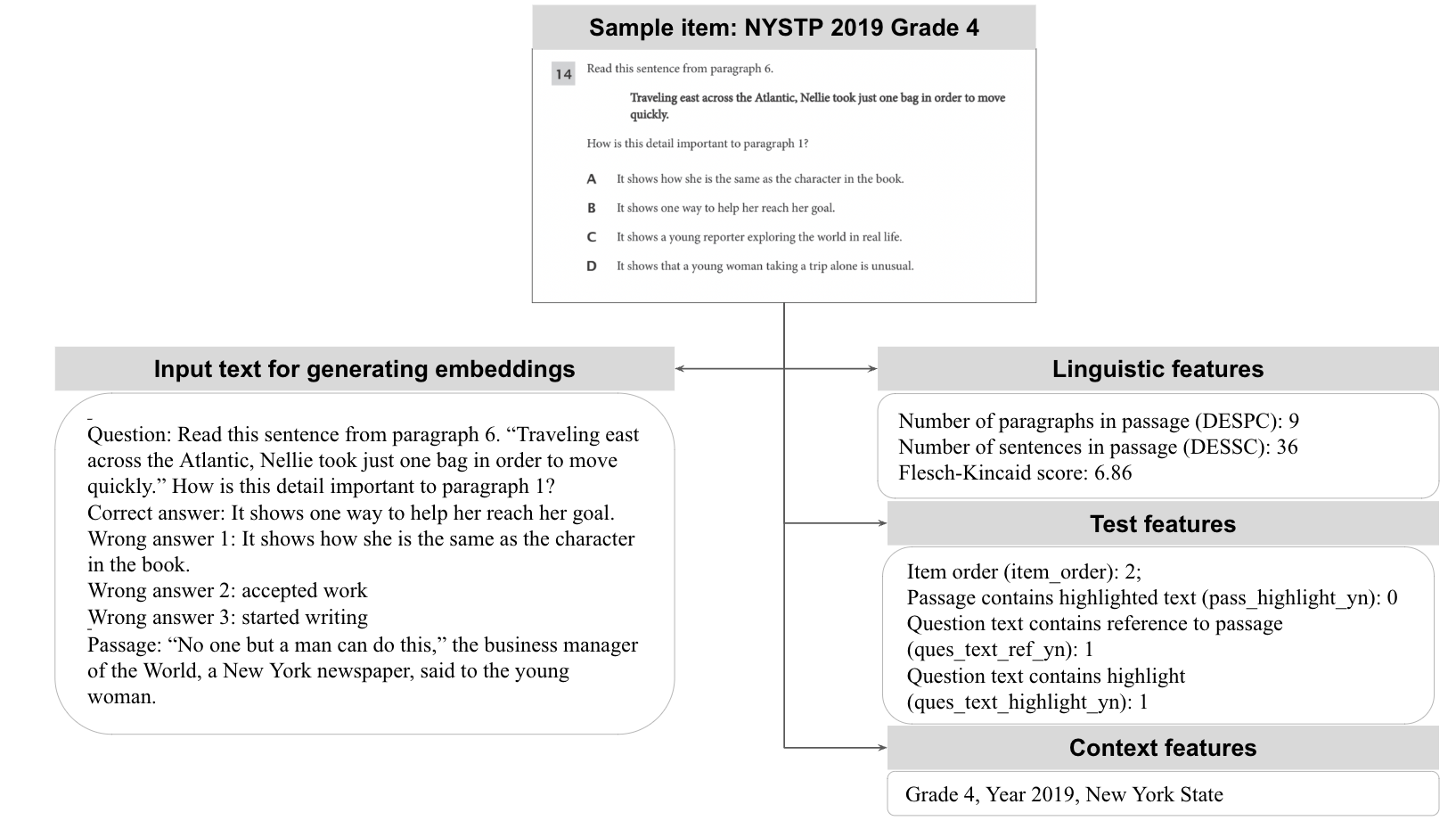}
\caption*   {Note. The figure shows a sample item from New York 2019 Grade 4 test. The box \textbf{"Input text for generating embeddings"} demonstrates how the passage and item is converted into a sentence. Embeddings are generated for this sentence. The boxes on the right show samples of annotated features for the item: \textbf{Linguistic features} shows some examples of text analysis of the reading passage. \textbf{Test features} show the relevant item design elements. \textbf{Context features} capture the item's year, state, and targeted grade. }
\end{figure}

\subsubsection{Context Features}
Respondent context characteristics are expected to be some of the most important predictors of difficulty. In this dataset, the context variables are the state, year, and grade of respondents, for which average difficulty values are reported. 

\subsubsection{Test Features}
Test features refer to the presentation or design elements in a test which could influence response. For the passage, this includes the use of highlighted (bolded, italicized, or underlined) text to emphasize important elements. For questions, this refers to the use of highlighted text (e.g., the use of bold text in ``6. Which sentence \textbf{best} states the main idea of `Around the World'?''), or inclusion of relevant text or sentences from the reading passage. For example, in the following question, the sentence in bold is repeated from the passage: ``Read this sentence from paragraph 6. \textbf{`Traveling east across the Atlantic, Nellie took just one bag to move quickly.'} How is this detail important to paragraph 1?''. The inclusion of these test characteristics is captured as indicator variables (=1 when given test feature is present in the item). Table \ref{tab:assesFeatures} lists the variables that capture relevant test characteristics about an item.

. 

\begin{table}[!htb]
\caption{Variables: Passage and Item Characteristics}
\label{tab:assesFeatures}
\centering
\resizebox{\textwidth}{!}{%
\begin{tabular}{ll}
\hline
\textbf{Variable} & \textbf{Description} \\ \hline
\texttt{item\_order} & Number indicating the order in which item appears after a passage  \\ 
\texttt{pass\_highlight\_yn}  & Indicator (1/0) if the passage text is highlighted \\ 
\texttt{ques\_text\_ref\_yn} & Indicator (1/0) if the question text includes relevant paragraphs or sentences from text \\
\texttt{ques\_text\_highlight\_yn} & Indicator (1/0) if the question text includes text in bold or underlined \\\hline
\end{tabular}%
}
\end{table}

\subsubsection{Text Analysis Features}
Measures of linguistic complexity of reading passages are generated using Coh-Metrix software package developed included as predictors for the analysis. The indicators are divided into the following categories: (1) Descriptive metrics, such as count of paragraphs, count of sentences, mean number of words in a sentence, and number of letters or syllables in a word (2) Readability metrics, such as Flesch-Kincaid score and Coh-Metrix L2 Readability (3) Ease of reading or text ``easability'', estimated through principal components analysis (PCA) components, measuring syntactic simplicity, vocabulary concreteness, referential cohesion, and deep cohesion (4) Text cohesion, where more cohesive texts are easier to read; measured by word overlap across text or semantic overlap in text (5) Deep cohesion in text, measured by use of causal verbs like \textit{although} and \textit{overall}, (6) Syntactic complexity, measured by use of more complex syntax such as density of noun phrases or syntax similarity of complex sentences (7) Word complexity, based on word frequency or psychological traits of words such as age of acquisition

These measures are generated by the Coh-Metrix software \citep{mcnamara2014automated}, except for the Flesch-Kincaid score (generated by readability package in Python\footnote{https://pypi.org/project/readability/}), and count of paragraphs in passage and word count of passage, which was generated in R.  

\subsection{Sentence Embedding from Large Language Models}
LLM-generated embeddings for question text and distractors are used as predictors in difficulty modeling. Distractors are tagged with \texttt{correct answer} or \texttt{wrong answer} as appropriate. The question text, correct answer and distractors, and passage are merged into one input string, for which embeddings are generated (Figure \ref{EmbeddingGen}).   We experiment with BERT \citep{DBLP:journals/corr/abs-1810-04805}, ModernBERT base model with 149M parameters \citep{modernbert} and LLaMA-3.1-8B \citep{dubey2024llama} for generation of sentence embedding. 

To generate embeddings from ModernBERT\footnote{https://huggingface.co/answerdotai/ModernBERT-base} and BERT\footnote{https://huggingface.co/google-bert/bert-base-cased}, each item is converted to a statement by merging the question text and available options as follows: (1) question text (2) a tag for \texttt{correct answer}, followed by the correct option (3) a tag for \texttt{wrong answer}, followed by the first wrong answer in the question. This is repeated for the remaining wrong answers (4) Reading passages. The objective is to capture interactions between the passage, question text, and distractors. The BERT tokenizer is used to convert these sentences into token IDs. The embeddings are derived from the last hidden state of the model. Specifically, for each sentence, we determine the position of the last meaningful token by identifying the index of the final non-padding token. The hidden state corresponding to this last token is extracted and used as the sentence-level embedding. The embeddings are averaged over the sentence length to get constant-sized tensors (length 768 for BERT and ModernBERT, length 4096 for LlAMA) for each sentence. The dimensionality of these embeddings is also compressed using PCA to capture 80\% of variation. PCA on embeddings also makes it impossible to recover the question text and options from the embeddings, which would be a test security concern for high-stakes tests. 

We also experimented with other methods of generating embeddings from BERT by removing the reading passage from the text input to tokenizer. We also generate cosine similarity between \texttt{correct answer} and \texttt{wrong answers} as new predictor variables. The results have some sensitivity to the method of embedding generation, but main conclusions do not change (Appendix \ref{Appendix B}, Table \ref{tab:appBERTemb}).

\subsection{Prediction Model}
Item-level difficulty (i.e., $b_j$ from Eqn \ref{eqn:diff}) is predicted through a Ridge ($L_2$) penalized regression that tuned the regularization penalty parameter ($\lambda$)\footnote{Elastic net models that tune the ratio between ridge and lasso ($\alpha$) always converged to the more parsimonius ridge model} via 5-fold cross-validation (repeated and averaged 5 times with a 80\% train, 20\% test split). All predictor variables are centered at 0 with standard deviation set to 1. The performance of the prediction models is measured by two indicators: RMSE, and Pearson Correlation Coefficient between the true ($y_j$) and predicted outcome ($\widehat{y_j}$) variable \citep{alkhuzaey2023text}.

\section{Results}
The predictors include the following categories of variables: (1) context characteristics of state, grade, and year (2) test characteristics (3) linguistic features of reading passage and item (4) LLM-generated embeddings (either the full set of embeddings or PCA components that capture 80\% of the variation). The baseline model for comparison sets the predicted difficulty to the mean difficulty for the item; this baseline RMSE was 0.92. This serves as a comparison point for model performance as measured by RMSE, which are dependent on the sample data set.

\begin{table}[tb]
\caption{Results for Predicting Item Difficulty}
\label{tab:mainresults}
\centering
\resizebox{\textwidth}{!}{
\begin{tabular}{lcccc}
\toprule
 & \multicolumn{2}{c}{\textbf{RMSE}} & \multicolumn{2}{c}{\textbf{Correlation}} \\ 
 \cmidrule(lr){2-3} \cmidrule(lr){4-5}
& \textbf{Train}   & \textbf{Test}   & \textbf{Train}     & \textbf{Test}     \\ \midrule
\multicolumn{5}{l}{\textbf{Results from human annotated features}} \\  
State, Grade, Year & 0.60 & 0.64 & 0.74 & 0.72\\
Test features & 0.89 & 0.92 & 0.13 & 0.05\\ 
Text analysis features & 0.59 & 0.62 & 0.75 & 0.75\\
All annotated features and context & 0.53 & 0.59 & 0.81 & 0.76 \\\midrule
\multicolumn{5}{l}{\textbf{Results from LLM embeddings only}} \\  
BERT embeddings & 0.60 & 0.66 & 0.76 & 0.71\\
LlAMA embeddings & 0.44 & 0.66 & 0.89 & 0.70\\
ModernBERT embeddings & 0.58 & 0.62 & 0.77 & 0.76\\ \midrule
\multicolumn{5}{l}{\textbf{Results from LLM embeddings and annotated features}} \\  
Annotated features \& BERT embeddings & 0.60 & 0.64 & 0.76 & 0.73\\
Annotated features \& PCA on BERT embeddings & 0.58 & 0.64 & 0.76 & 0.72\\
Annotated features \& LlAMA embeddings & 0.47 & 0.62 & 0.87 & 0.75\\
Annotated features \& PCA on LlAMA embeddings & 0.59 & 0.63 & 0.75 & 0.73\\
Annotated features \& ModernBERT embeddings &  0.57 & 0.61 & 0.78 & 0.76\\
Annotated features \& PCA on ModernBERT embeddings & 0.59 & 0.63 & 0.75 & 0.73\\\midrule
\multicolumn{5}{l}{\textbf{Results from LLM embeddings, annotated features, and context}} \\  
All features \& BERT embeddings & 0.59 & 0.64 & 0.77 & 0.74\\
All features \& PCA on BERT embeddings & 0.52 & 0.60 & 0.81 & 0.76\\
All features \& LlAMA embeddings & 0.47 & 0.61 & 0.87 & 0.76\\
All features \& PCA on LlAMA embeddings & 0.52 & 0.59 & 0.81 & 0.76\\
All features \& ModernBERT embeddings & 0.57 & 0.61 & 0.78 & 0.77\\
All features \& PCA on ModernBERT embeddings & 0.52 & 0.59 & 0.81 & 0.77\\

\bottomrule
\end{tabular}
}
\caption*{Note. Outcome variable is easiness drawn from IRT scale based on NWEA MAP Reading scale Spring 2020}
\end{table}

\subsection{Prediction results from annotated features and embeddings}
The best performing model uses all available features and PCA on embeddings from ModernBERT or LlAMA (Table \ref{tab:mainresults}). The best RMSE on test data is 0.59 and correlation is 0.77.\footnote{Appendix \ref{Appendix B} reports the results when other vertical scales are used to convert p-values to IRT easiness: for different scales, RMSE varies between 0.53 and 0.61, and correlation varies between 0.75 and 0.85. The RMSE is 3.24 and correlation is 0.96 when different scales are used for pre and post 2020 years. This is not strictly comparable with other results which use the same scales for all years.}. It is not surprising that the best model uses all available features for prediction. We also note that PCA on embeddings performs marginally better than actual embeddings; this is likely because of the small dataset size compared to the number of predictors. 

\subsection{Prediction results from LLM embeddings only}
Results from only LLM embeddings, without using any respondent context characteristics or linguistic analysis of text, are close to the best model (Table \ref{tab:mainresults}. The correlation of true and predicted difficulty is 0.76 for ModernBERT with an RMSE of 0.62. When all additional features are added, model performance increases marginally for RMSE to 0.61 and correlation to 0.77. BERT and LlAMA show slightly bigger improvements in model performance when all features are added: RMSE decreases from 0.66 when only BERT embeddings are included to 0.64 when all features are included; the correlation between true and predicted difficulty in the test dataset increases from 0.71 to 0.74. For LlAMA, RMSE decreases from 0.66 to 0.61, and correlation increases from 0.70 to 0.76.


\subsection{Prediction results from annotated features}
Model performance for annotated features (context, test, and text analysis) is also close to the best model. This is largely because of the text analysis features, which are able to predict item difficulty. This suggests that LLMs and text analysis features might be capturing similar information. Text analysis features offer the added benefit of being interpretable by humans and need less computation power, which might also be an advantage over using LLMs.  

The overall prediction power of test features by themselves is low as would be expected. However, these features have some of the largest and significant coefficient values in the analysis; while these coefficients cannot be interpreted in an elastic net model, this suggests that the variables capture important information about item difficulty. 


It is also important to note that these features can predict item difficulty without including context characteristics (state, grade, year). These models could hence be used to predict average item difficulty before extensive field pilots, potentially reducing pilot costs. 



\section{Discussion}

This paper aims to predict item difficulty using its text content. This approach bridges the gap between manual item creation and modern computational methods, providing insights into the nuanced interplay between text complexity and student response behaviors. The study underscores the transformative potential of transformer-based models, which can generate meaningful representations of items without fine-tuning or additional context information. By rescaling difficulties using IRT and vertical ability scales, the study addressed the challenge of comparing average difficulty across grades, offering a scalable solution when there is no common population or anchor items for traditional test equating and linking approaches. This item difficulty prediction model could be useful for stakeholders like teachers or testing firms, who would be interested in creating items with desirable item properties like difficulty. For instance, an item generated through a generative AI model could be assigned a difficulty score using this model; this could then be used to assemble appropriate tests or worksheets. 

Our model demonstrated strong performance in estimating item difficulty, achieving a correlation coefficient of 0.77 between predicted and observed difficulty for models that incorporated LLM embeddings and all available item features. To the best of our knowledge, this correlation between true and predicted difficulty is one of the best for these tasks. The RMSE is high, but we expect that it is a function of the constraints of the dataset; the dataset we use is relatively small and the outcome variable is average item difficulty converted to a logit scale. The best model also reduces RMSE from a baseline value of 0.92 to 0.59. Other work that predicts difficulty as a continuous measure reports Pearson's correlation between 0.38 and 0.4 \citep{huang2017question} and a reduction in RMSE from 23.65 to 22.45 \citep{ha2019predicting}. 

The best results were obtained for the models that combined text analysis and human generated features with LLM embeddings. Surprisingly, in our dataset, using LLM-based embeddings or text analysis features was close to the best performing model. These findings suggest that LLMs could predict item difficulty without additional linguistic analysis or human annotations. Embeddings were generated from three models, BERT, ModernBERT and LLaMA; the results are also similar for the three LLM models. LLaMA embeddings were computationally more expensive to generate but their model performance was better. ModernBERT achieved the best performance, suggesting it might be best suited to such an item difficulty modeling task. BERT's strong performance despite being the simplest model could be a function of the limited size of the dataset, where variations in embedding generation might not lead to improvements in results.  


While the findings affirm the utility of automated methods, they also highlight several limitations of static embeddings in fully capturing dynamic student interactions. First, the reliance on state-specific datasets from New York and Texas may limit the generalizability of the findings to other educational contexts or student populations. Second, the embeddings generated from LLMs are dependent on pre-trained models that may not be optimally tuned to K-12 educational texts or standardized test items, potentially missing domain-specific nuances. Lastly, the model’s predictions are based on aggregated item difficulty, which may oversimplify individual differences in student abilities or test-taking contexts.


Future research should focus on expanding the dataset to include diverse state tests and international benchmarks, such as PISA or data high-stakes tests, to enhance the robustness and applicability of the model. Additionally, fine-tuning LLMs on domain-specific corpora could further improve their ability to capture educational nuances. A comprehensive evaluation of item difficulty prediction should explore its impact on downstream tasks, such as automated item generation and field testing, to validate the practical utility of the proposed model in educational settings. Last, but perhaps most important, this model could be used by teachers or test creators to get early feedback on item properties.


\clearpage

\printbibliography
\clearpage

\begin{appendices}  
\renewcommand\thefigure{\thesection.\arabic{figure}}    
\setcounter{figure}{0}

\renewcommand\thetable{\thesection.\arabic{table}}  
\setcounter{table}{0}

\section{Variables used as predictors}
\label{Appendix A}

\subsection{Linguistic features}
\setcounter{table}{0}

\begin{longtable}{p{3cm}p{14cm}}
    \caption{Table of Measures and Descriptions}
    \label{tab:measures} \\
    \hline
    \multicolumn{1}{p{3cm}}{ \textbf{Measure}} & 
    \multicolumn{1}{p{10cm}}{ \textbf{Description}} \\ \hline
    \multicolumn{2}{l}{\textbf{Descriptive metrics}} \\ \hline
    PassageWordCount & Word count of the passage \\ 
    DESPC & Paragraph count, number of paragraphs \\ 
    DESSC & Sentence count, number of sentences \\ 
    DESPL & Paragraph length, number of sentences in a paragraph, mean \\ 
    DESPLd & Paragraph length, number of sentences in a paragraph, standard deviation \\ 
    DESSL & Sentence length, number of words, mean \\ 
    DESSLd & Sentence length, number of words, standard deviation \\ 
    DESWLsy & Word length, number of syllables, mean \\ 
    DESWLsyd & Word length, number of syllables, standard deviation \\ 
    DESWLlt & Word length, number of letters, mean \\ 
    DESWLltd & Word length, number of letters, standard deviation \\ \hline
    \multicolumn{2}{l}{\textbf{Readability metrics}} \\ \hline
    FK & Flesch-Kincaid score \\
    RDFRE & Flesch-Kincaid	Reading	Ease\\
    RDFKGL  & Flesch-Kincaid	Grade level	\\
    RDL2 & Coh-Metrix	L2	Readability\\\hline
    \multicolumn{2}{p{17cm}}{\textbf{Text reading ``easability''}: Principal Components analysis mapped to dimensions of reading ease i.e., syntactic simplicity, word concreteness, referential cohesion, deep cohesion, verb cohesion, connectivity, and temporality} \\ \hline
    PCNARz & Text Easability PC Narrativity, z score (e.g., tells a story) \\ 
    PCNARp & Text Easability PC Narrativity, percentile \\ 
    PCSYNz & Text Easability PC Syntactic simplicity, z score (e.g., sentences contain fewer words and simpler sentences) \\ 
    PCSYNp & Text Easability PC Syntactic simplicity, percentile \\ 
    PCCNCz & Text Easability PC Word concreteness, z score (e.g., more concrete words)  \\ 
    PCCNCp & Text Easability PC Word concreteness, percentile \\ 
    PCREFz & Text Easability PC Referential cohesion, z score (e.g., words or ideas repeat across text) \\ 
    PCREFp & Text Easability PC Referential cohesion, percentile \\ 
    PCDCz & Text Easability PC Deep cohesion, z score (e.g., text contains causal and intentional connectives that aid with comprehension)  \\ 
    PCDCp & Text Easability PC Deep cohesion, percentile \\ 
    PCVERBz & Text Easability PC Verb cohesion, z score (e.g., overlapping verbs in text) \\ 
    PCVERBp & Text Easability PC Verb cohesion, percentile \\ 
    PCCONNz & Text Easability PC Connectivity, z score (e.g., contains explicit adversative, additive, and comparative connectives) \\ 
    PCCONNp & Text Easability PC Connectivity, percentile \\ 
    PCTEMPz & Text Easability PC Temporality, z score (e.g., text has cues about temporality such as tense and aspect) \\ 
    PCTEMPp & Text Easability PC Temporality, percentile \\\hline
    \multicolumn{2}{l}{\textbf{Referential cohesion:} Word overlap across text, measured by indices of word and sentence overlaps} \\ \hline 
    CRFNO1 & Noun overlap, adjacent sentences, binary, mean \\ 
    CRFAO1 & Argument overlap, adjacent sentences, binary, mean \\ 
    CRFSO1 & Stem overlap, adjacent sentences, binary, mean \\ 
    CRFNOa & Noun overlap, all sentences, binary, mean \\ 
    CRFAOa & Argument overlap, all sentences, binary, mean \\ 
    CRFSOa & Stem overlap, all sentences, binary, mean \\ 
    CRFCWO1 & Content word overlap, adjacent sentences, proportional, mean \\ 
    CRFCWO1d & Content word overlap, adjacent sentences, proportional, standard deviation \\ 
    CRFCWOa & Content word overlap, all sentences, proportional, mean \\ 
    CRFCWOad & Content word overlap, all sentences, proportional, standard deviation \\ \hline
    \multicolumn{2}{p{17cm}}{\textbf{Referential cohesion:} Semantic overlap in text measured by Latent Semantic Analysis} \\ \hline
    LSASS1 & LSA overlap, adjacent sentences, mean \\ 
    LSASS1d & LSA overlap, adjacent sentences, standard deviation \\ 
    LSASSp & LSA overlap, all sentences in paragraph, mean \\ 
    LSASSpd & LSA overlap, all sentences in paragraph, standard deviation \\ 
    LSAPP1 & LSA overlap, adjacent paragraphs, mean \\ 
    LSAPP1d & LSA overlap, adjacent paragraphs, standard deviation \\ 
    LSAGN & LSA given/new, sentences, mean \\ 
    LSAGNd & LSA given/new, sentences, standard deviation \\ \hline
    \multicolumn{2}{p{18cm}}{\textbf{Lexical diversity}: Variety of words in a given text. Lexical diversity is lower and cohesion is higher when words are repeated across text} \\ \hline
    LDTTRc & Lexical diversity, type-token ratio, content word lemmas \\ 
    LDTTRa & Lexical diversity, type-token ratio, all words \\ 
    LDMTLD & Lexical diversity, MTLD, all words \\ 
    LDVOCD & Lexical diversity, VOCD, all words \\\hline
    \multicolumn{2}{p{17cm}}{\textbf{Connectives}: Incidence of connectives per 1000 words, which reflects cohesive links between ideas and text organization. Cohesion is higher when there is greater incidence of connectives. } \\ \hline 
    CNCAll & All connectives incidence  \\ 
    CNCCaus & Causal connectives incidence (e.g., \textit{because, so}) \\ 
    CNCLogic & Logical connectives incidence (e.g., \textit{and, or}) \\ 
    CNCADC & Adversative and contrastive connectives incidence (e.g., \textit{although, whereas}) \\ 
    CNCTemp & Temporal connectives incidence (e.g., \textit{first, until}) \\ 
    CNCTempx & Expanded temporal connectives incidence \\ 
    CNCAdd & Additive connectives incidence (e.g., \textit{and, moreover})\\ 
    CNCPos & Positive connectives incidence (e.g., \textit{also, moreover}) \\ 
    CNCNeg & Negative connectives incidence \\ 
    \\\hline
    \multicolumn{2}{p{17cm}}{\textbf{Situation model:} Representations of deeper meaning in text using Lexical Semanic Analysis and WordNet} \\ \hline 
    SMCAUSv & Causal verb incidence (e.g., \textit{although, whereas})\\ 
    SMCAUSvp & Causal verbs and causal particles incidence \\ 
    SMINTEp & Intentional verbs incidence \\ 
    SMCAUSr & Ratio of causal particles to causal verbs \\ 
    SMINTEr & Ratio of intentional particles to intentional verbs \\ 
    SMCAUSlsa & LSA verb overlap \\ 
    SMCAUSwn & WordNet verb overlap \\ 
    SMTEMP & Temporal cohesion, tense and aspect repetition, mean \\ 
    \\\hline
    \multicolumn{2}{p{17cm}}{\textbf{Syntactic complexity:} Includes measures of complex syntax in text and Syntactic Pattern Density; where higher density reflects more informational dense text with more complex syntax} \\ \hline 
    SYNLE & Left embeddedness, words before main verb, mean \\ 
    SYNNP & Number of modifiers per noun phrase, mean \\
    SYNMEDpos & Minimal Edit Distance, part of speech \\ 
    SYNMEDwrd & Minimal Edit Distance, all words \\ 
    SYNMEDlem & Minimal Edit Distance, lemmas \\ 
    SYNSTRUTa & Sentence syntax similarity, adjacent sentences, mean \\ 
    SYNSTRUTt & Sentence syntax similarity, all combinations, across paragraphs, mean \\\hline
    DRNP & Noun phrase density, incidence \\ 
    DRVP & Verb phrase density, incidence \\ 
    DRAP & Adverbial phrase density, incidence \\ 
    DRPP & Preposition phrase density, incidence \\ 
    DRPVAL & Agentless passive voice density, incidence \\ 
    DRNEG & Negation density, incidence \\ 
    DRGERUND & Gerund density, incidence \\ 
    DRINF & Infinitive density, incidence \\ \hline
    \multicolumn{2}{p{17cm}}{\textbf{Word complexity}: information about parts of speech (e.g., relative frequency of types of word categories such as nouns, verbs and adverbs, word frequency (e.g., frequency of words used in text in CELEX database), and word complexity based on psychological ratings (e.g., age of acquisition of words, word concreteness)}  \\ \hline
    WRDNOUN & Noun incidence \\ 
    WRDVERB & Verb incidence \\ 
    WRDADJ & Adjective incidence \\ 
    WRDADV & Adverb incidence \\ 
    WRDPRO & Pronoun incidence \\ 
    WRDPRP1s & First person singular pronoun incidence \\ 
    WRDPRP1p & First person plural pronoun incidence \\ 
    WRDPRP2 & Second person pronoun incidence \\ 
    WRDPRP3s & Third person singular pronoun incidence \\ 
    WRDPRP3p & Third person plural pronoun incidence \\
    WRDFRQc & word frequency for content words using CELEX database, mean \\ 
    WRDFRQa & CELEX Log frequency for all words, mean \\ 
    WRDFRQmc & CELEX Log minimum frequency for content words, mean \\
    WRDAOAc & Age of acquisition for content words, mean \\ 
    WRDFAMc & Familiarity for content words, mean \\ 
    WRDCNCc & Concreteness for content words, mean \\ 
    WRDIMGc & Imagability for content words, mean \\ 
    WRDMEAc & Meaningfulness, Colorado norms, content words, mean \\ 
    WRDPOLc & Polysemy for content words, mean \\ 
    WRDHYPn & Hypernymy for nouns, mean \\ 
    WRDHYPv & Hypernymy for verbs, mean \\ 
    WRDHYPnv & Hypernymy for nouns and verbs, mean \\ \hline
\end{longtable}

\newpage

\setcounter{table}{0}
\section{Robustness checks}
\label{Appendix B}

\subsection{Robustness to choice of grade-level growth scale}
This section examines the robustness of results to selecting an alternate vertical scale to convert p-values to IRT difficulty. Table \ref{tab:nwea_staar_appdx} describes alternate scales that could be used: NWEA 2015 MAP reading achievement standard (reported separately for literary texts and informational text), NWEA 2020 reading achievement as measured in Fall or Winter,  Texas STAAR 2023-24 performance standards (``approaches grade-level'', ``meets grade level'', and ``masters grade level'' performance), and Texas STAAR 2017-18 readiness standard. These different scales were selected to account for differences in state contexts, as well as differences in grade level scales before and after 2020.

\begin{figure}[!htb]
    \centering
\caption{\textbf{Robustness to Vertical Scale Choice}}
    \label{testCor}    \includegraphics[width=\textwidth]{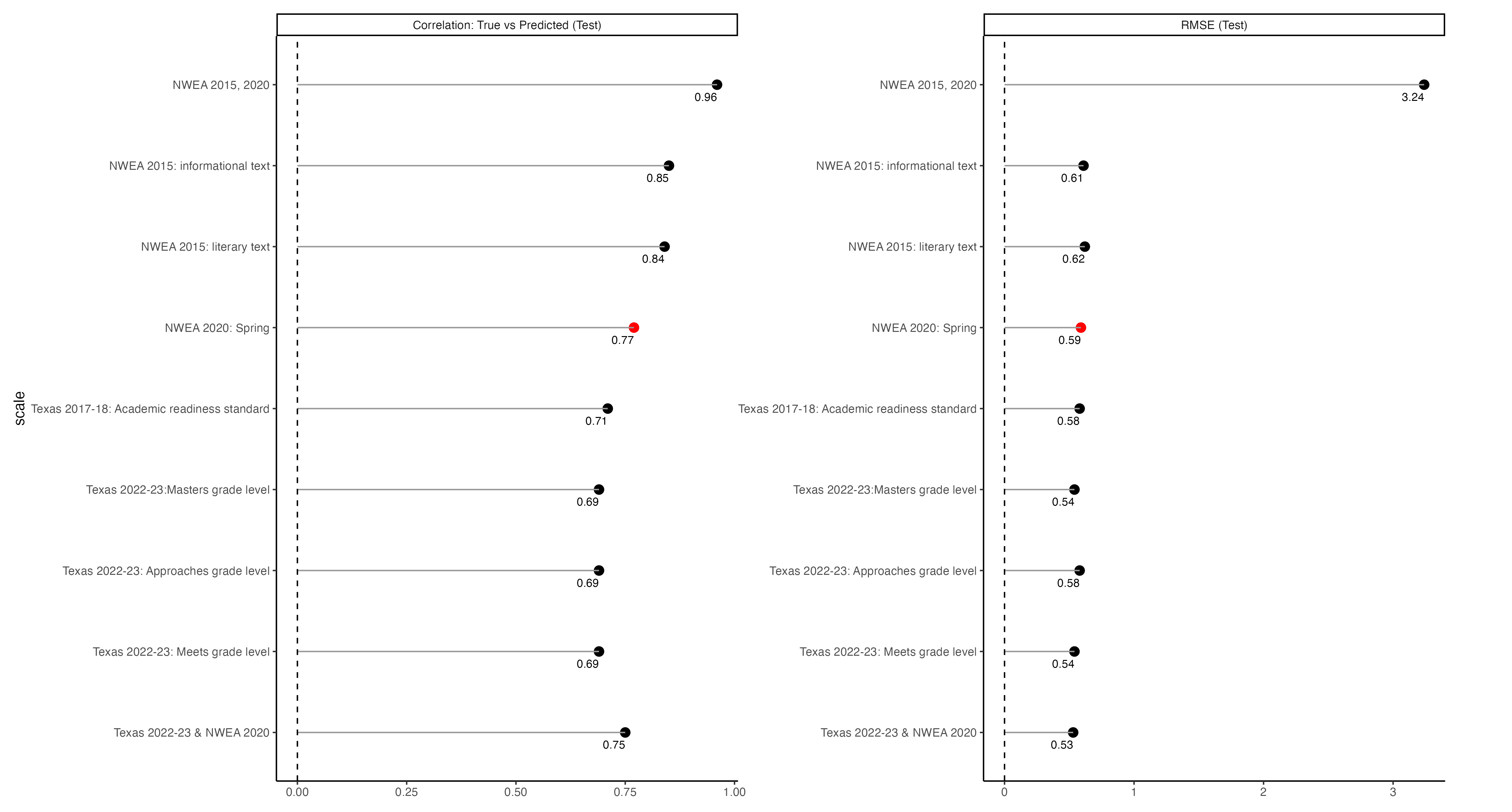}
    \caption*{ Note. (Left panel) Correlation of true vs predicted difficulty for test data; (Right panel) RMSE for true vs predicted difficulty. The main results reported in the paper use the NWEA Spring 2020 vertical scale, shown in red in the figure. Note that results reported in the paper are conservative and not extremely sensitive to the choice of vertical scale.} 
\end{figure}

Results for test RMSE, as well as correlation between true and predicted RMSE are shown in Figure \ref{testCor}. We note that the results presented in the main section are conservative and do not fluctuate drastically with choice of scale. The greatest variation is observed when we use NWEA Informational Text 2015 scale for 2018 and 2019, and the NWEA 2020 Spring scale for 2021 to 2023. The RMSE increases to 3.24, likely due to the introduction of variation across years. However, correlation between true and predicted difficulty also increases to 0.96. However, since it is difficult to obtain precise estimates of the change in vertical grade level scales before and after 2020, these results are not highlighted in the main paper.

\begin{table}[!htb]
    \centering
    \caption{Alternate Vertical Growth Scales: NWEA MAP 2015, NWEA MAP 2020 and Texas Meets Grade Level Performance 2023-24}
    \resizebox{\textwidth}{!}{
        \begin{tabular}{ccc c ccc c ccc c c}
        \hline
        \multirow{2}{*}{Grades} & \multicolumn{2}{c}{NWEA 2015} && \multicolumn{3}{c}{NWEA 2020} && \multicolumn{3}{c}{Texas STAAR 2023-24} && \multicolumn{1}{c}{Texas 2017-18} \\ \cline{2-3} \cline{5-7} \cline{9-11} \cline{13-13}
        & \shortstack{Literary\\ Text} & \shortstack{Informational\\ Text} & & Fall & Winter & Spring & & \shortstack{Approaches\\grade-level} & \shortstack{Meets\\grade-level} & \shortstack{Masters\\grade-level} & & \shortstack{Readiness\\standard}    \\ \hline
        Grade 3 & 192.4 & 191.6 & & 186.62 & 195.91 & 200.74 & & 1345 & 1467 & 1596 && 1386 \\ 
        Grade 4 & 201.2 & 200.7 & & 196.67 & 202.5 &  204.83 & & 1414 & 1552 & 1663 && 1473 \\ 
        Grade 5 & 207.9 & 207.4 & & 204.48 & 210.19 & 210.98 & & 1471 & 1592 & 1700 && 1508\\ 
        Grade 6 & 212.3 & 212.1 & & 210.17 & 213.81 & 215.36 & & 1535 & 1634 & 1749 && 1554\\ 
        Grade 7 & 216.3 & 216.1 & & 214.2 & 217.09 & 216.81 & & 1564 & 1669 & 1771 && 1603\\ 
        Grade 8 & 220.0 & 220.0 & & 218.9 & 220.52 & 220.93 & & 1592 & 1698 & 1803 && 1625\\ \hline
        \end{tabular}
     }
    \label{tab:nwea_staar_appdx}
\end{table}

\begin{figure}[htb!]
\caption{\textbf{Robustness to scales: [Left] Correlation of true vs predicted difficulty [Right] RMSE of true vs predicted difficulty}}
    \centering
    \begin{subfigure}{0.49\textwidth}
        \centering
        \includegraphics[width=\linewidth]{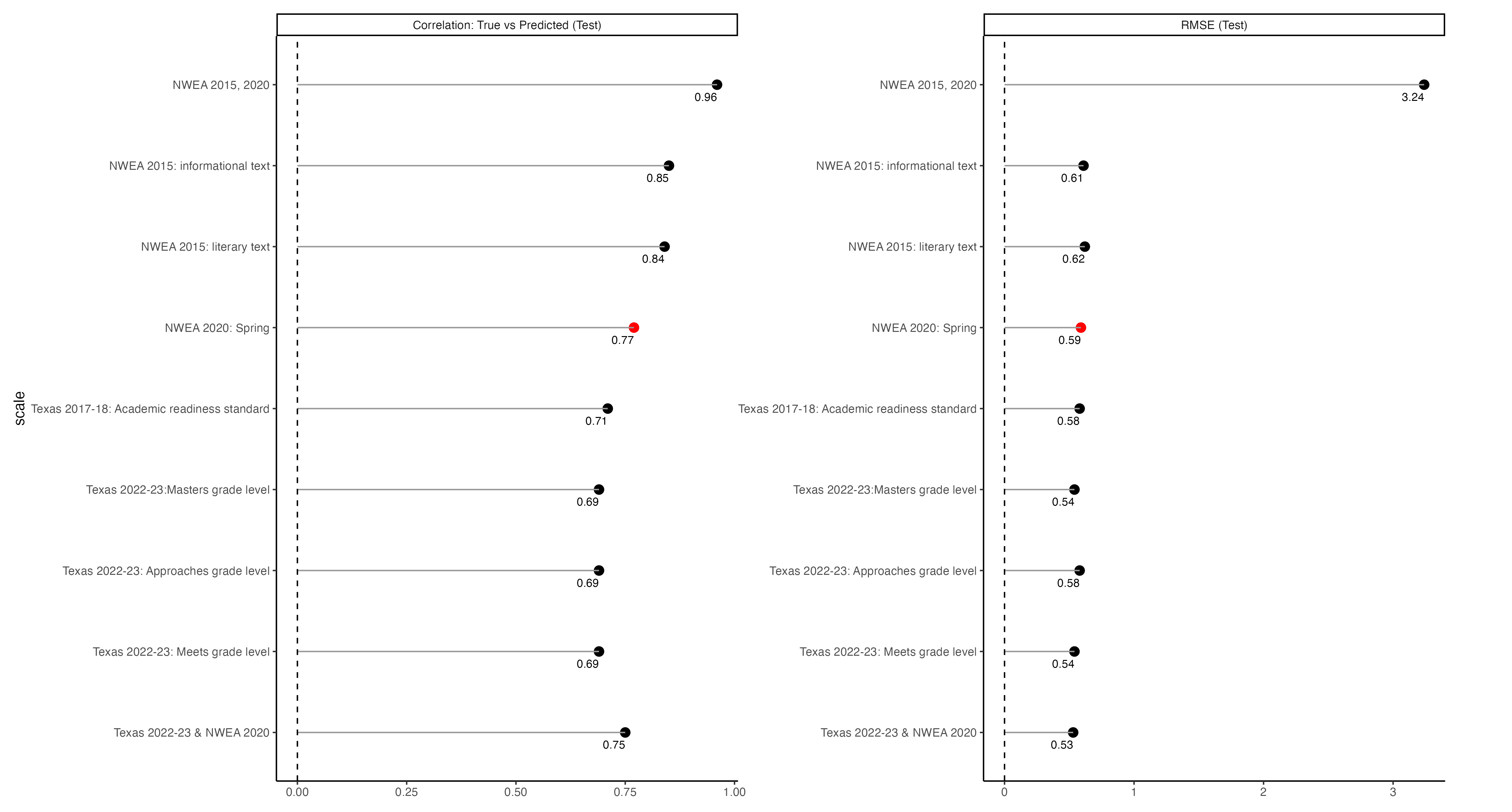}
        \caption{Correlation of true vs predicted difficulty (b)}
        \label{fig:correlation}
    \end{subfigure}
    \hfill
    \begin{subfigure}{0.49\textwidth}
        \centering
        \includegraphics[width=\linewidth]{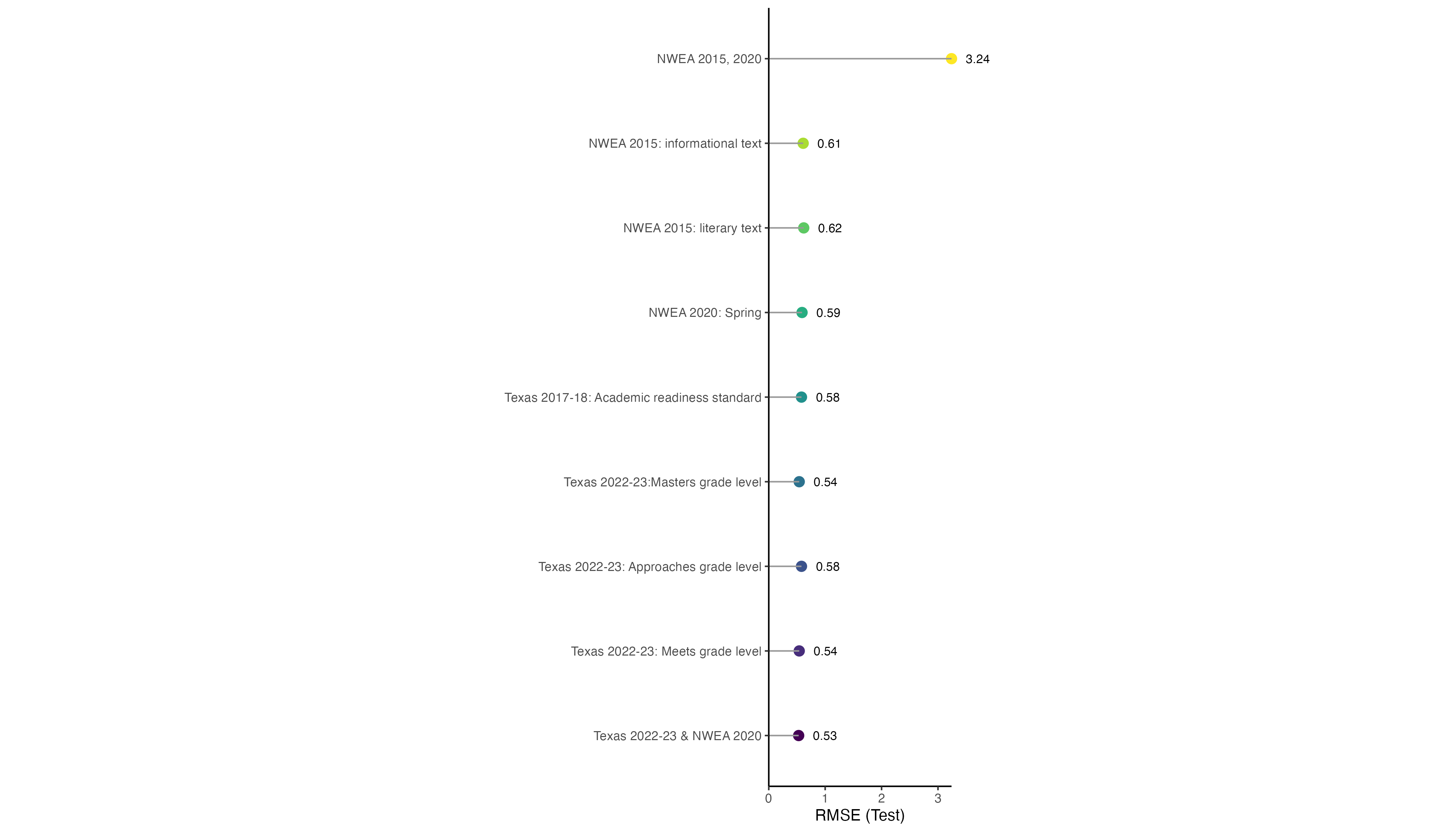}
        \caption{RMSE of true vs predicted difficulty (b)}
        \label{fig:rmse}
    \end{subfigure}
    \caption{\textbf{Robustness to scales: Correlation and RMSE for test data}}
    \label{fig:robustness}
\end{figure}

\subsection{BERT embeddings: Robustness to text input format}
\begin{table}[!ht]
\caption{BERT model: Robustness of Prediction Results}
\label{tab:appBERTemb}
\centering
\resizebox{\textwidth}{!}{
\begin{tabular}{lcccc}
\toprule
 & \multicolumn{2}{c}{\textbf{RMSE}} & \multicolumn{2}{c}{\textbf{Correlation}} \\ 
 \cmidrule(lr){2-3} \cmidrule(lr){4-5}
& \textbf{Train}   & \textbf{Test}   & \textbf{Train}     & \textbf{Test}     \\ \midrule
\multicolumn{5}{l}{\textbf{Main Results: Text input is Question + Correct answer + Wrong answers + Passage}} \\  
\midrule
LLM embeddings: BERT & 0.60 & 0.66 & 0.76 & 0.71\\
Assessment characteristics, text analysis metrics, \& BERT embeddings & 0.60 & 0.64 & 0.76 & 0.73\\
Assessment characteristics, text analysis metrics, \& PCA on BERT embeddings & 0.58 & 0.64 & 0.76 & 0.72\\ \midrule
\multicolumn{5}{l}{\textbf{Alternate Results 1: Text input is Question + Correct answer + Wrong answers}} \\  \hline
LLM embeddings: BERT & 0.65 & 0.72 & 0.71 & 0.63\\
Assessment characteristics, text analysis metrics, \& BERT embeddings & 0.60 & 0.65 & 0.76 & 0.73\\
Assessment characteristics, text analysis metrics, \& PCA on BERT embeddings & 0.58 & 0.61 & 0.76 & 0.75\\ \midrule
\multicolumn{5}{l}{\textbf{Alternate Results 2: Main results with Cosine Similarity Variables}} \\  \hline
LLM embeddings:BERT & 0.60 & 0.66 & 0.76 & 0.72\\
Assessment characteristics, text analysis metrics, BERT embeddings & 0.60 & 0.64 & 0.76 & 0.74\\
Assessment characteristics, text analysis metrics, PCA on BERT embeddings & 0.58 & 0.63 & 0.76 & 0.73\\ 
\bottomrule
\end{tabular}
 }
\end{table}
The main results report BERT embeddings generated for a statement that merges question text, correct answer, all wrong answers, and passage (Figure \ref{EmbeddingGen}). The reading passage was removed from the statement used to generate BERT embeddings for \texttt{Alternate Results 1} (Table \ref{tab:appBERTemb}). If the passage is relevant to how BERT embeddings capture information about the item, we would expect decline in model performance. We do observe this decline: RMSE increases from 0.66 for \texttt{Main Results} to 0.72 for \texttt{Alternate Results 1}, and correlation decreases from 0.72 to 0.63. However, prediction results are similar when human annotated features are included in the model, and when the model uses PCA on BERT embeddings instead of the full embeddings.  This suggests that human annotated features might contain complementary information that compensates model performance. This is also interesting because BERT embeddings are generated using a section of the passage (sentence length is truncated at 512 characters). This suggests that a segment of passage still includes useful information; hence, the main results use embeddings generated with the passage included in the text input to the tokenizer. 

We also include new predictor variables: cosine similarity is calculated between the embeddings for correct answer and each of the distractors. Embeddings are generated for the sentence that combines: (1) question text (2) a tag for correct/wrong answer, followed by the correct/wrong answer. Next, we calculate cosine similarity between the embeddings for the correct answer and each of the wrong answers, giving us three cosine similarity variables. These variables are included as predictors in the main model. These results are reported as \texttt{Alternate Results 2}. We note marginal improvement in performance between \texttt{Main Results} and \texttt{Alternate Results 2}, where RMSE decreases from 0.64 to 0.63, and correlation marginally improves from 0.72 to 0.73. This suggests that differences or similarities in BERT embeddings for correct answers and distractors don't explain item difficulty in the context of this dataset.

\clearpage
\section{Results for unadjusted p-values}
This section presents the results for unadjusted p-values i.e., the p-values reported in the dataset that have not been adjusted for grade-level differences. As discussed in the methods section, this measure of difficulty does not have an inherent meaning. The results show a lower RMSE, likely because grade-level variation has not been accounted for. However, the correlation is also low (in the range of 0.10-0.3).  

\begin{table}[!ht]
\caption{Results for Predicting Item Difficulty: Outcome Variable is Easiness from IRT scale}
\label{tab:results}
\centering
\resizebox{\textwidth}{!}{
\begin{tabular}{lcccc}
\toprule
 & \multicolumn{2}{c}{\textbf{RMSE}} & \multicolumn{2}{c}{\textbf{Correlation}} \\ 
 \cmidrule(lr){2-3} \cmidrule(lr){4-5}
& \textbf{Train}   & \textbf{Test}   & \textbf{Train}     & \textbf{Test}     \\ \midrule
\multicolumn{5}{l}{\textbf{Results from human annotated features}} \\  
State, Grade, Year & 0.12 & 0.12  & 0.28 & 0.29\\
Test features & 0.13 & 0.13 & 0.15 & 0.10\\
Text analysis features & 0.12 & 0.13  & 0.36 & 0.11\\
\\\midrule
\multicolumn{5}{l}{\textbf{Results from LLM embeddings only}} \\  
Only BERT embeddings & 0.11 & 0.12  & 0.57 & 0.30\\
Only LlAMA embeddings & 0.10 & 0.12  & 0.69 & 0.35\\
Only ModernBERT embeddings & & & & \\\midrule
\multicolumn{5}{l}{\textbf{Results from LLM embeddings and annotated features}} \\  
All features \& BERT embeddings & 0.10 & 0.12  & 0.60 & 0.32\\
All features \& PCA on BERT embeddings & 0.12 & 0.12  & 0.46 & 0.25\\
All features \& LlAMA embeddings & 0.10 & 0.12  & 0.72 & 0.35\\
All features \& PCA on LlAMA embeddings & 0.12 & 0.12  & 0.46 & 0.25\\
All features \& PCA on ModernBERT embeddings & 0.12 & 0.12  & 0.45 & 0.25\\
\bottomrule
\end{tabular}
}
\end{table}


\end{appendices}
\end{document}